\pdfoutput=1

\documentclass[11pt,letterpaper]{article}
\usepackage[english]{babel}
\usepackage{emnlp2016}
\usepackage{times}
\usepackage{latexsym}
\usepackage{graphicx}
\usepackage{multirow}
\usepackage{booktabs}
\usepackage{url}
\usepackage{paralist}
\usepackage{etoolbox}
\usepackage{xspace}
\usepackage{color}
\usepackage[normalem]{ulem}
\usepackage[font=small]{caption}
\usepackage{adjustbox}
\usepackage{subcaption}

\graphicspath{ {.} }

\def\emnlppaperid{958}

\emnlpfinalcopy

\belowdisplayskip \abovedisplayskip

\newlength{\sectionReduceTop}
\newlength{\sectionReduceBot}
\newlength{\subsectionReduceTop}
\newlength{\subsectionReduceBot}
\newlength{\abstractReduceTop}
\newlength{\abstractReduceBot}
\newlength{\captionReduceTop}
\newlength{\captionReduceBot}
\newlength{\listReduceTop}
\newlength{\listReduceBot}
\newlength{\subsubsectionReduceTop}
\newlength{\subsubsectionReduceBot}

\newlength{\eqnReduceTop}
\newlength{\eqnReduceBot}

\newlength{\horSkip}
\newlength{\verSkip}

\newlength{\figureHeight}
\newcommand{\tableScale}{0.95}

\newcommand{\tfdataset}{VTFQ\xspace}
\newcommand{\capqsim}{\textbf{\textsc{q-c sim}}\xspace}
\newcommand{\qqsim}{\textbf{\textsc{q-q' sim}}\xspace}

\newcommand{\agentours}{\textbf{\textsc{agent-ours}}\xspace}
\newcommand{\agentbaseline}{\textbf{\textsc{agent-baseline}}\xspace}

\newcommand{\entropy}{\textbf{\textsc{entropy}}\xspace}
\newcommand{\captionmodelscores}{\textbf{\textsc{q-gen score}}\xspace}

\newcommand{\rulebased}{\textbf{\textsc{rule-based}}\xspace}
\newcommand{\lstm}{\textbf{\textsc{lstm}}\xspace}
\newcommand{\vqamlp}{\textbf{\textsc{vqa-mlp}}\xspace}
\newcommand{\bow}{\textbf{\textsc{bow}}\xspace}
\newcommand{\avgwtov}{\textbf{\textsc{avg. w2v}}\xspace}
\newcommand{\lstmwtov}{\textbf{\textsc{lstm w2v}}\xspace}

\newcommand{\ie}{\emph{i.e.,}\xspace}
\newcommand{\eg}{\emph{e.g.,}\xspace}
\newcommand{\Eg}{\emph{E.g.,}\xspace}

\title{Question Relevance in VQA:\\Identifying Non-Visual And False-Premise Questions}

\author{Arijit Ray$^1$, Gordon Christie$^1$, Mohit Bansal$^2$, Dhruv Batra$^{3,1}$, Devi Parikh$^{3,1}$ \\
$^1$Virginia Tech \quad $^2$UNC Chapel Hill\quad $^3$Georgia Institute of Technology \\ {\tt \{ray93,gordonac,dbatra,parikh\}@vt.edu} \\ {\tt mbansal@cs.unc.edu}}

\begin{document}

\maketitle

\begin{abstract}
Visual Question Answering (VQA) is the task of answering natural-language questions about images. We introduce the novel problem of determining the \emph{relevance of questions to images} in VQA. Current VQA models do not reason about whether a question is even related to the given image (\eg~\emph{What is the capital of Argentina?}) or if it requires information from external resources to answer correctly. This can break the continuity of a dialogue in human-machine interaction. Our approaches for determining relevance are composed of two stages. Given an image and a question, (1) we first determine whether the question is visual or not, (2) if visual, we determine whether the question is relevant to the given image or not. Our approaches, based on LSTM-RNNs, VQA model uncertainty, and caption-question similarity, are able to outperform strong baselines on both relevance tasks. We also present human studies showing that VQA models augmented with such question relevance reasoning are perceived as more intelligent, reasonable, and human-like. 
\end{abstract}

{
\setdefaultleftmargin{0pt}{}{}{}{}{}

\vspace{\sectionReduceTop}
\section{Introduction}
\vspace{\sectionReduceBot}
\label{sec:introduction}

Visual Question Answering (VQA) is the task of predicting a suitable answer given an image and a question about it. 
VQA models (\eg  \cite{antol2015vqa,ren2015exploring})  are typically discriminative models that take in image and question representations and output one of a set of possible answers.

Our work is motivated by the following key observation -- all current VQA systems always output an answer \emph{regardless of whether the input question makes any sense for the given image or not}. 
Fig.~\ref{fig:teaser} shows examples of relevant and irrelevant 
questions. 
When VQA systems are fed irrelevant questions as input, they understandably produce nonsensical answers (Q: \emph{``What is the capital of Argentina?''} A: \emph{``fire hydrant''}). Humans, on the other hand, are unlikely to provide such nonsensical answers and will instead answer that this is irrelevant or use another knowledge source to answer correctly, when possible. 
We argue that this implicit assumption by all VQA systems -- that an input question is always relevant for the input image -- is simply untenable as VQA systems move beyond standard academic datasets to interacting with real users, who may be unfamiliar, or malicious.
The goal of this work is to make VQA systems more human-like by providing them the capability to identify relevant questions.

\begin{figure}[t]
\centering
\includegraphics[width=0.969\columnwidth]{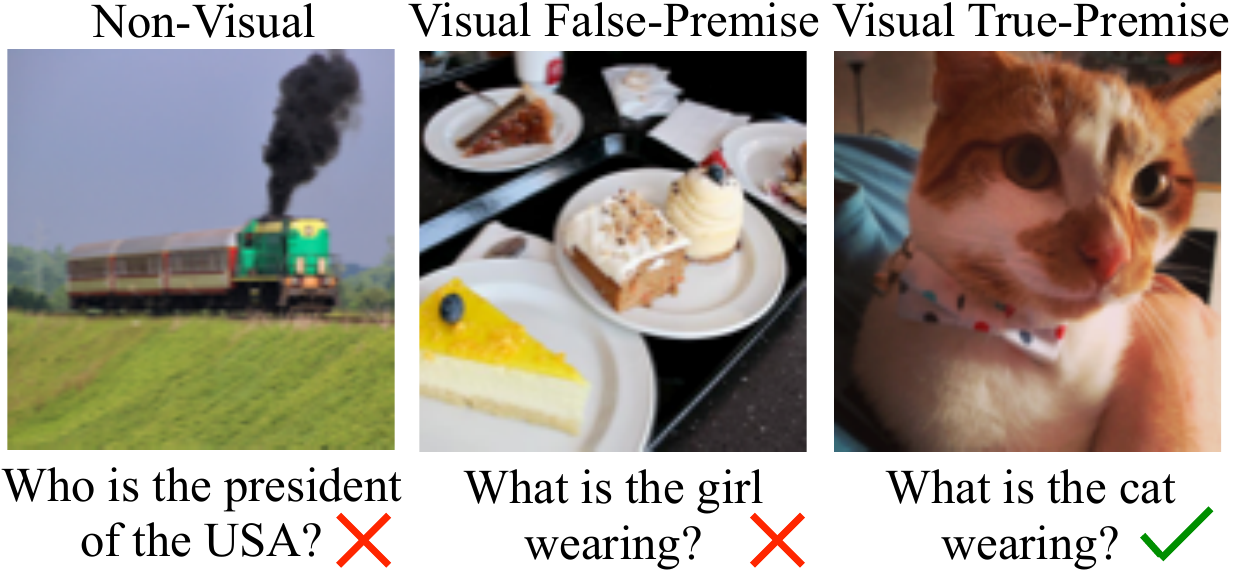}
\caption{
Example irrelevant (non-visual,  false-premise) and relevant (visual true-premise) questions in VQA.
}
\label{fig:teaser}
\vspace{\captionReduceBot}
\vspace{-10pt}
\end{figure}

While existing work has reasoned about cross-modal similarity,
being able to identify whether a question is relevant to a given image is a novel problem with real-world applications. In human-robot interaction, being able to identify questions that are dissociated from the perception data available is important. The robot must decide whether to process the scene it perceives or query external world knowledge resources to provide a response.

As shown in Fig.~\ref{fig:teaser}, we study three types of question-image pairs:
\textbf{Non-Visual}. These questions are not questions about images at all -- they do not require information from 
\emph{any}
image to be answered
(\eg \emph{``What is the capital of Argentina?''}).
\textbf{Visual False-Premise}. 
While visual, these questions do not apply to the given image. For instance, the question \emph{``What is the girl wearing?''} makes sense only for images that contain a girl in them.
\textbf{Visual True-Premise}. These questions are relevant to (\ie have a premise which is true) the image at hand.

We introduce datasets and train models to recognize both non-visual and false-premise 
question-image (QI) pairs
in the context of VQA. First, we identify whether a question is visual or non-visual; if visual, we identify whether the question has a true-premise for the given image. 
For visual vs.~non-visual question detection, we use a Long Short-Term Memory (LSTM) recurrent neural network (RNN) trained on part of speech (POS) tags to capture visual-specific linguistic structure. For true vs.~false-premise question detection, we present one set of approaches that use the uncertainty of a VQA model, and another set that use pre-trained captioning models to generate relevant captions (or questions) for the given image and then compare them to the given question to determine relevance.

Our proposed models achieve accuracies of $92\%$ for detecting non-visual, and $74\%$ for detecting false-premise questions, which significantly outperform strong baselines. 
We also show through human studies that a VQA system that reasons about question relevance is picked \emph{significantly more often} as being more intelligent, human-like and reasonable than a baseline VQA system which does not. Our code and datasets are publicly available on the authors' webpages.

\vspace{\sectionReduceTop}
\section{Related Work}
\vspace{\sectionReduceBot}
\label{sec:related_work}

There is a large body of existing work that reasons about cross-modal similarity: how well an image matches a query tag~\cite{liu2009boost} in text-based image retrieval, how well an image matches a caption~\cite{feng2013automatic,xu2015show,NIPS2011_4470,karpathy2015deep,Fang_2015_CVPR}, and how well a video matches a description~\cite{donahue2015long,lin2014visual}.

In our work, if a question is deemed irrelevant, the VQA model says so, as opposed to answering the question anyway. This is related to perception systems that do not respond to an input where the system is likely to fail. Such failure prediction systems have been explored in vision ~\cite{zhang2014predicting,devarakota2007confidence} and speech~\cite{zhao2012automatic,sarma2004context,choularton2009early,voll2008improving}. 
Others attempt to provide the most meaningful answer instead of suppressing the output of a model that is expected to fail for a given input.
One idea is to avoid a highly specific prediction if there is a chance of being wrong, and instead make a more generic prediction that is more likely to be right~\cite{deng2012hedging}.
\newcite{malinowski2014multi} use semantic segmentations in their approach to question answering, where they reason that objects not present in the segmentations should not be part of the answer.

To the best of our knowledge, our work is the first to study the relevance of questions in VQA. 
\newcite{chen2014understandingwww} classify users' intention of questions for community question answering services. 
Most related to our work is \newcite{visualtext}. They extract visual text from within Flickr photo captions to be used as supervisory signals for training image captioning systems. Our motivation is to endow VQA systems the ability to detect non-visual questions to respond in a human-like fashion. Moreover, we also detect a more fine-grained notion of question relevance via true- and false-premise. 

\vspace{\sectionReduceTop}
\section{Datasets}
\vspace{\sectionReduceBot}
\label{sec:datasets}

For the task of detecting visual vs.~non-visual questions, we assume all questions in the VQA dataset~\cite{antol2015vqa} are visual,
since the Amazon Mechanical Turk (AMT) workers were specifically instructed to ask questions about a displayed image while creating it. 
We also collected non-visual philosophical and general knowledge questions from the internet (see appendix). 
Combining the two, we have 121,512 visual questions from the validation set of VQA and 9,952\footnote{High accuracies on this task in our experiments indicate that this suffices to learn the corresponding linguistic structure.} generic non-visual questions collected from the internet. We call this dataset Visual vs.~Non-Visual Questions (VNQ). 

We also collect a dataset of true- vs.~false-premise questions by showing AMT workers images paired with random questions from the VQA dataset and asking them to annotate whether they are applicable or not. We had three workers annotate each QI pair. We take the majority vote as the final ground truth label.\footnote{78\% of the time all three votes agree.} We have 10,793 QI pairs on 1,500 unique images out of which $79\%$ are non-applicable (false-premise). 
We refer to this visual true- vs.~false-premise questions dataset as \tfdataset.

Since there is a class imbalance in both of these datasets, we report the average per-class (\ie normalized) accuracy for all approaches. All datasets are publicly available.

\vspace{\sectionReduceTop}
\section{Approach}
\vspace{\sectionReduceBot}
\label{sec:approach}

\begin{table*}[ht!]
\vspace{\captionReduceTop}
\centering
\setlength{\tabcolsep}{11pt}
\scalebox{\tableScale}{
\begin{tabular}{ccccccc}
\toprule
\multicolumn{2}{c}{\textbf{Visual vs. Non-Visual}} & \multicolumn{5}{c}{\textbf{True- vs. False-Premise}} \\
\cmidrule[0.75pt](l){1-2}
\cmidrule[0.75pt](l){3-7}
\rulebased & \lstm & \entropy 
& 
\vqamlp
& 
\captionmodelscores 
& 
\capqsim
& 
\qqsim
\\ 
\midrule
75.68 & \textbf{92.27} & 59.66 & 64.19 & 57.41 & 74.48 & \textbf{74.58} \\
\bottomrule
\end{tabular}
}
\caption{Normalized accuracy results (averaged over 40 random train/test splits) for visual vs. non-visual detection and true- vs. false-premise detection. \rulebased and \captionmodelscores were not averaged because they are deterministic.}
\label{results}
\end{table*}

Here we present our approaches for detecting (1) visual vs.~non-visual QI pairs, and (2) true- vs. false-premise QI pairs.

\vspace{\subsectionReduceTop}
\subsection{Visual vs. Non-Visual Detection}
\vspace{\subsectionReduceBot}

Recall that the task here is to detect visual questions from non-visual ones. Non-visual questions, such as 
\emph{``Do dogs fly?''} or \emph{``Who is the president of the USA?''}, often tend to have a difference in the linguistic structure from that of visual questions, such as \emph{``Does this bird fly?''} or \emph{``What is this man doing?''}. 
We compare our approach (\lstm) with a baseline (\rulebased):

\begin{compactenum}
\item \rulebased. A rule-based approach to detect non-visual questions based on the part of speech (POS)\footnote{We use spaCy POS tagger \cite{honnibalimproved}.}
tags and dependencies of the words in the question. \Eg if a question has a plural noun with no determiner before it and  is followed by a singular verb (\emph{``Do dogs fly?''}), it is a non-visual question.\footnote{See appendix for examples of such hand-crafted rules.}
\item \lstm. 
We train an LSTM with 100-dim hidden vectors to embed the question into a vector and predict visual vs. not. 
Instead of feeding question words ([`what', `is', `the', `man', `doing', `?']), the input to our LSTM is embeddings of POS tags of the words ([`pronoun', `verb', `determiner', `noun', `verb']). Embeddings of the POS tags are learnt end-to-end. This captures the structure of image-grounded questions, rather than visual vs. non-visual topics. The latter are less likely to generalize across domains.

\end{compactenum}

\vspace{\subsectionReduceTop}
\subsection{True- vs. False-Premise Detection}
\vspace{\subsectionReduceBot}

Our second task is to detect whether a question Q entails 
a false-premise for an image I. We present two families of approaches to measure this QI `compatibility': 
(i) using uncertainty in VQA models, and (ii) using pre-trained captioning models.

\paragraph{Using VQA Uncertainty.}

Here we work with the hypothesis that 
if a VQA model is uncertain 
about the answer to
a QI pair, the question 
may
be irrelevant for the given image since the uncertainty 
may mean
it has not seen similar QI pairs in the training data. 
We test two approaches:
\begin{compactenum}
\item \entropy. We compute the entropy of the softmax output from a state-of-the art VQA model 
\cite{antol2015vqa,Lu2015}
for a given QI pair and train a 
three-layer multilayer perceptron (MLP) on top with 3 nodes in the hidden layer.
\item \vqamlp. We feed in the softmax output to a three-layer MLP with 100 nodes in the hidden layer, and train it as a binary classifier to predict whether a question has a true- or false-premise for the given image.
\end{compactenum}

\paragraph{Using Pre-trained Captioning Models.}

Here we utilize (a) an image captioning model, and 
(b) an image question-generation model -- to measure QI compatibility. 
Note that both these models generate natural language 
capturing the semantics of an image -- one in the form of statement, 
the other in the form of a question. 
Our hypothesis is that a given question is relevant to the given image if it is similar to the language generated by these models for that image. Specifically: 

\begin{compactenum}
\item Question-Caption Similarity (\capqsim). 
We use NeuralTalk2~\cite{karpathy2015deep} pre-trained on the MSCOCO dataset~\cite{lin2014microsoft} (images and associated captions) to generate a caption C 
for the given image, and then compute a learned similarity 
between Q and C (details below).

\item Question-Question Similarity (\qqsim). 
We use NeuralTalk2 re-trained (from scratch) on the questions 
in the VQA dataset to generate a question Q' for the image. 
Then, we compute a learned similarity between Q and Q'. 
\end{compactenum}
We now describe our learned Q-C similarity function (the Q-Q' similarity is 
analogous). Our Q-C similarity model is a 2-channel LSTM+MLP (one channel 
for Q, another for C). 
Each channel sequentially reads word2vec embeddings of the 
corresponding language via an LSTM. 
The last hidden state vectors (40-dim) from the 2 LSTMs are concatenated 
and fed as inputs to the MLP, 
which outputs a 2-class (relevant vs. not) softmax. The entire model is 
learned end-to-end on the \tfdataset dataset. We also experimented with 
other representations (\eg~bag of words) for Q, Q', C, 
which are included in the appendix for completeness.

Finally, we also compare our proposed models above to a simpler baseline (\captionmodelscores), where 
we compute the probability of the input question Q under the learned question-generation model. 
The intuition here is that since the question generation model has 
been trained only on relevant questions (from the VQA dataset), 
it will assign a high probability to Q if it is relevant. 

\vspace{\sectionReduceTop}
\section{Experiments and Results}
\vspace{\sectionReduceBot}
\label{sec:experiments_results}

The results for both experiments are presented in Table~\ref{results}. We present results averaged over 40 random train/test splits. \rulebased and \captionmodelscores were not averaged because they are deterministic.

\paragraph{Visual vs. Non-Visual Detection.} 
We use a random set of 100,000 questions from the VNQ dataset for training, and the remaining 31,464 for testing.
We see that \lstm performs 16.59\% (21.92\% relative) better than \rulebased.

\paragraph{True- vs. False-Premise  Detection.}
We use a random set of 7,195 ($67\%$)
QI pairs from the \tfdataset dataset to train and the remaining 3,597 ($33\%$)
to test. 
While the VQA model uncertainty based approaches (\entropy, \vqamlp) 
perform reasonably well (with the MLP helping over raw entropy), the 
learned similarity 
approaches perform much better ($10.39\%$ gain in normalized accuracy). 
High uncertainty of the model may suggest that a similar QI pair 
was not seen during training; however, that does not 
seem to translate to detecting irrelevance. 
The language generation models (\capqsim, \qqsim) seem 
to work significantly better at modeling 
the semantic interaction between the question and the image. 
The generative approach (\captionmodelscores) is outperformed by the discriminative approaches (\vqamlp, \capqsim, \qqsim) that are trained explicitly for the task at hand.
We show qualitative examples of \qqsim for true- vs. false-premise detection in Fig.~\ref{fig:qual_main}.

\vspace{\sectionReduceTop}
\section{Human Qualitative Evaluation}
\vspace{\sectionReduceBot}
\label{sec:human_studies}

\begin{figure*}[ht!]
	\centering
	\begin{subfigure}[b]{0.24\textwidth}
		\includegraphics[width=\textwidth]{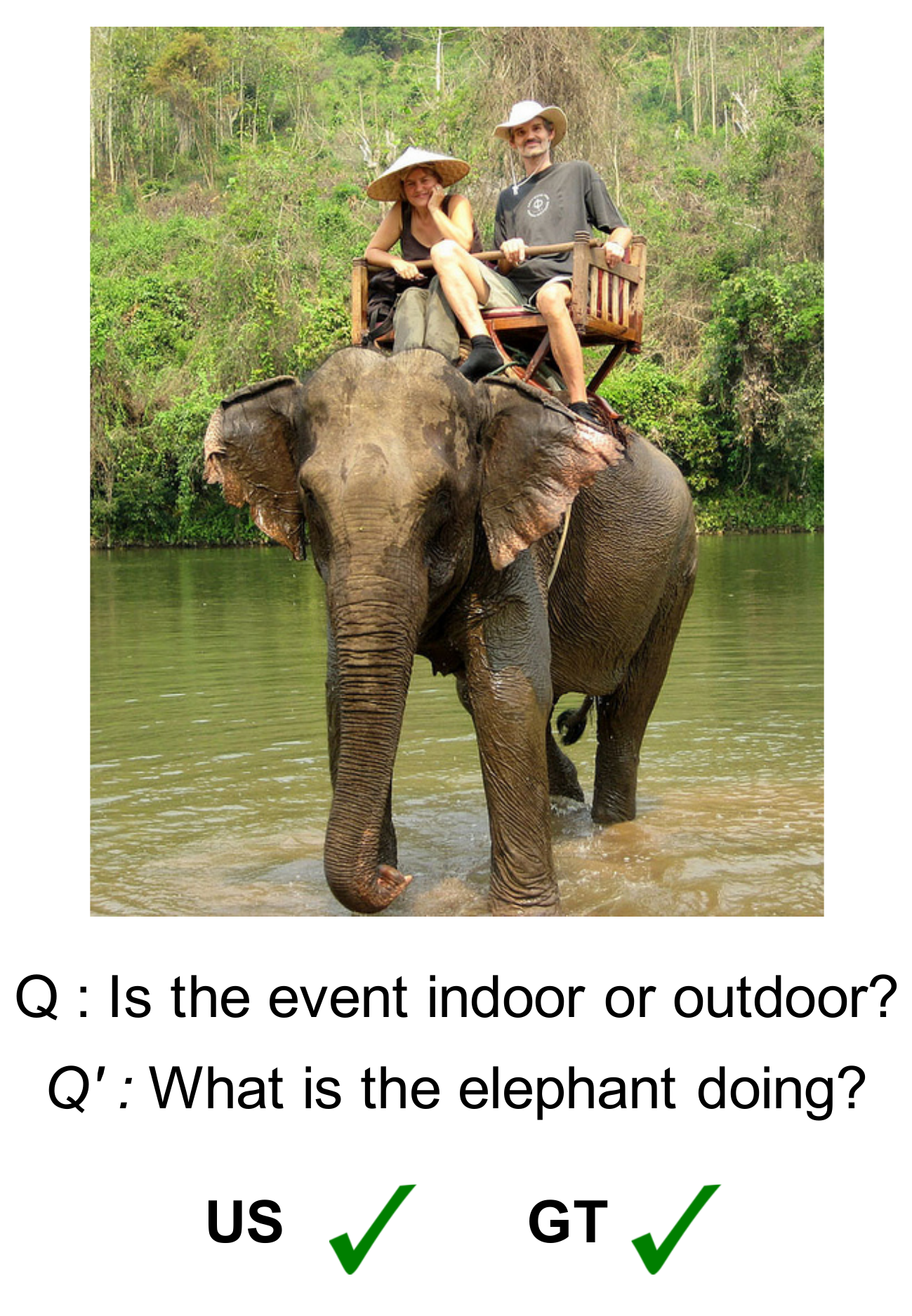}
		\caption{}
		\label{fig:q1}
	\end{subfigure}
    \begin{subfigure}[b]{0.24\textwidth}
		\includegraphics[width=\textwidth]{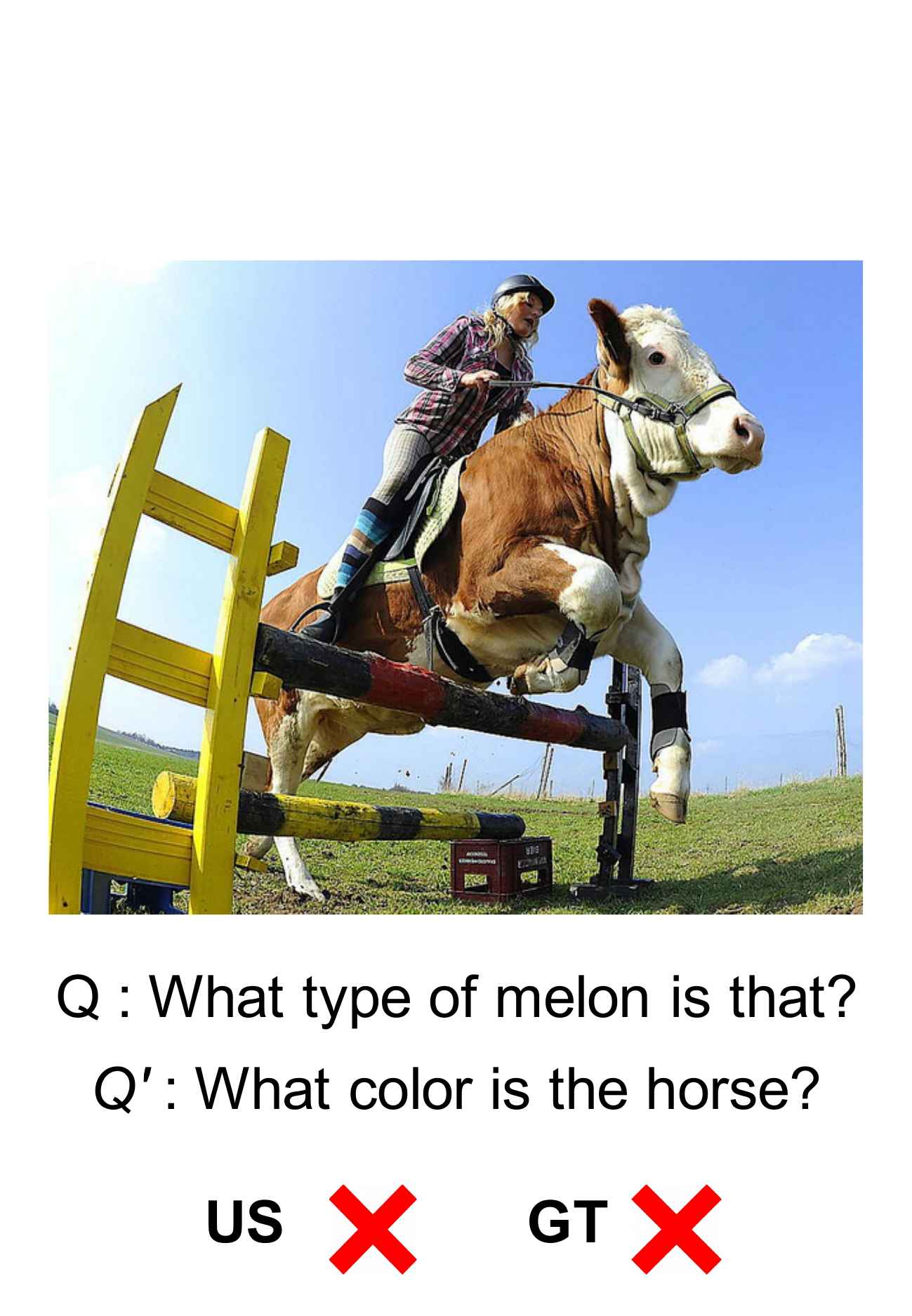}
		\caption{}
		\label{fig:q1}
	\end{subfigure}
    \begin{subfigure}[b]{0.24\textwidth}
		\includegraphics[width=\textwidth]{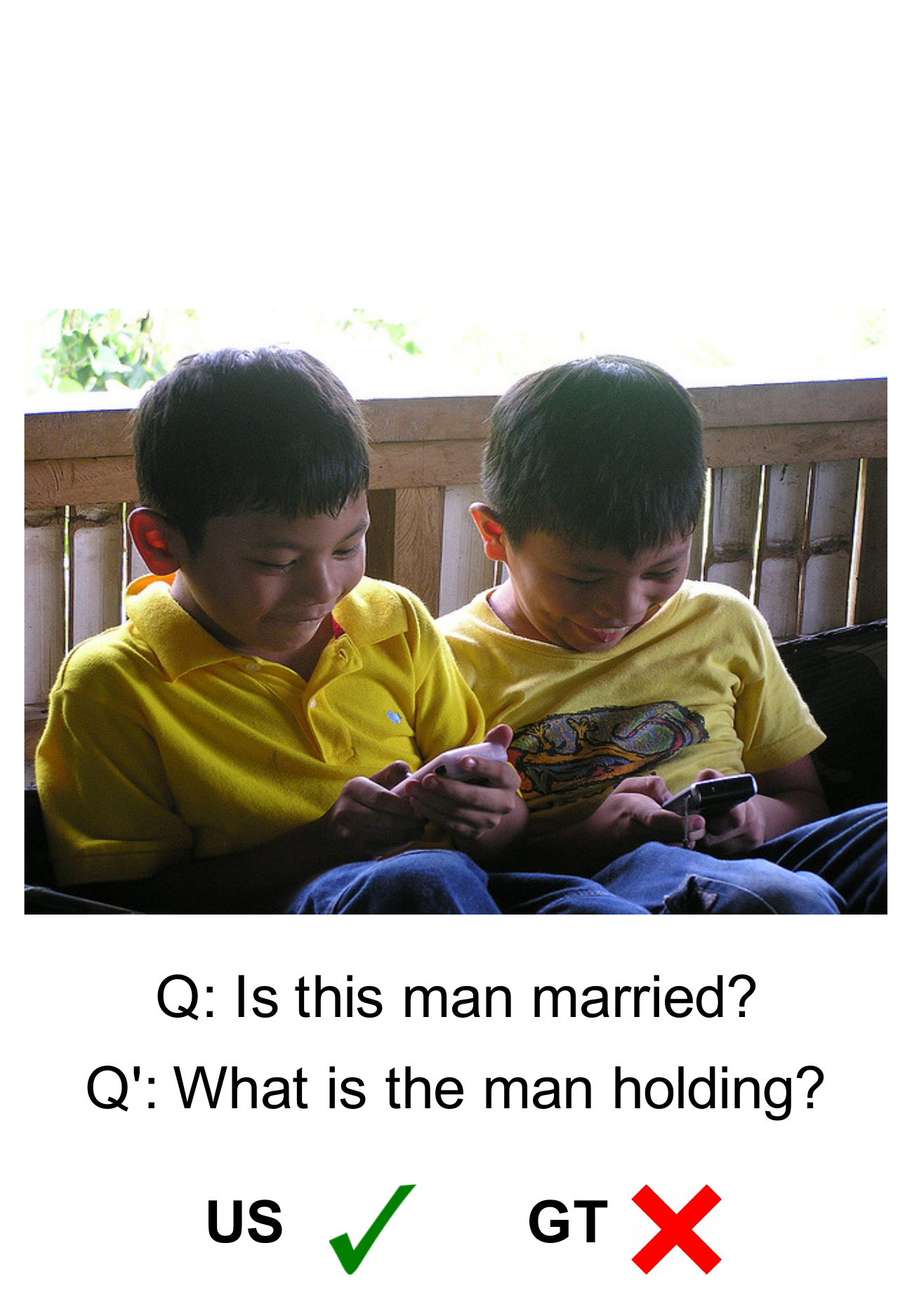}
		\caption{}
		\label{fig:q1}
	\end{subfigure}
    \begin{subfigure}[b]{0.24\textwidth}
		\includegraphics[width=\textwidth]{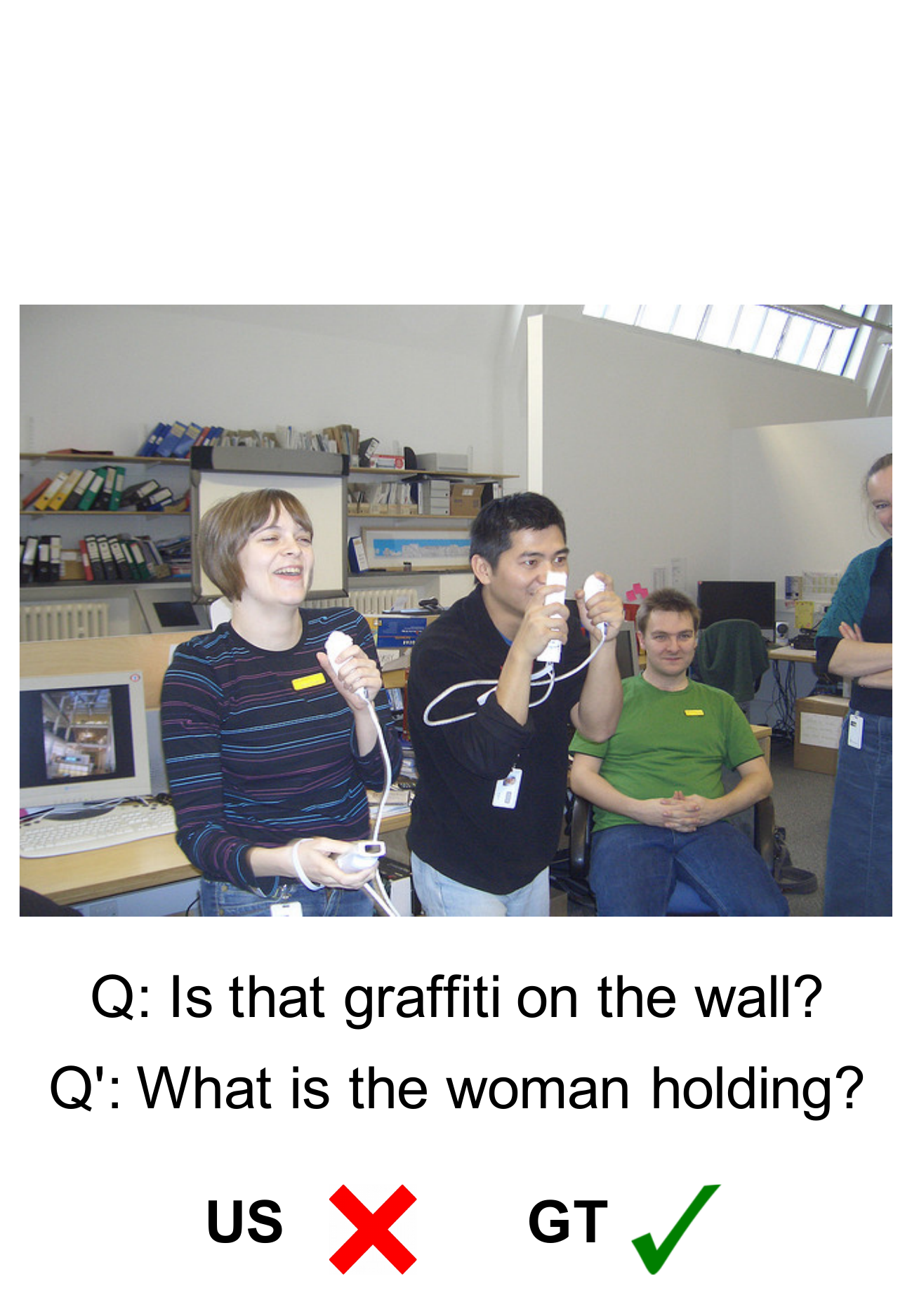}
		\caption{}
		\label{fig:q1}
	\end{subfigure}
	\caption{Qualitative examples for \qqsim. (a) and (b) show success cases, and (c) and (d) show failure cases. Our model predicts true-premise in (a) and (c), and false-premise in (b) and (d). In all examples we show the original question Q and the generated question Q'.}
	\label{fig:qual_main}
\end{figure*}

We also perform human studies where we compare two agents:
(1) \agentbaseline -- always answers every question. 
(2) \agentours -- reasons about question relevance before responding. 
If question is classified as visual true-premise, 
\agentours answers the question using the same VQA model as \agentbaseline (using~\cite{Lu2015}). Otherwise, it responds with a prompt indicating that the question does not seem meaningful for the image.

A total of 120 questions 
(18.33\% relevant,  81.67\% irrelevant, mimicking the distribution of the VTFQ dataset) 
were used. Of the relevant questions, 54\% were answered correctly by the VQA model. Human subjects on AMT were shown the response of both agents and  asked to pick the agent that sounded more intelligent, more reasonable, and more human-like 
after every observed QI pair. Each QI pair was assessed by 5 different subjects. Not all pairs were rated by the same 5 subjects. In total, 28 unique AMT workers participated in the study.

\agentours was picked 65.8\% of the time as the winner, \agentbaseline was picked only 1.6\% of the time, and both considered equally (un)reasonable in the remaining cases. 
We also measure the percentage of times  each robot gets picked by the workers for true-premise, false-premise, and non-visual questions. These percentages are shown in Table~\ref{tab:percent_picked}.

\begin{table}[h!]
\vspace{\captionReduceTop}
\centering
\scalebox{\tableScale}{
\begin{tabular}{cccc}
\toprule
& 
\textbf{\begin{tabular}[c]{@{}c@{}}True-\\ Premise\end{tabular}} & 
\textbf{\begin{tabular}[c]{@{}c@{}}False-\\ Premise\end{tabular}} & 
\textbf{\begin{tabular}[c]{@{}c@{}}Non-\\ Visual\end{tabular}} \\
\midrule
\agentours & 22.7 & 78.2 & 65.0 \\
\agentbaseline & 04.7 & 01.4 & 00.0 \\
\textbf{Both} & 27.2 & 03.8 & 10.0 \\
\textbf{None} & 45.4 & 16.6 & 25.0 \\
\bottomrule
\end{tabular}
}
\caption{Percentage of times each robot gets picked by AMT workers as being more intelligent, more reasonable, and more human-like for true-premise, false-premise, and non-visual questions.}
\label{tab:percent_picked}
\end{table}

Interestingly, humans often prefer \agentours over \agentbaseline 
even when \emph{both models are wrong} -- 
\agentbaseline answers the question incorrectly and 
\agentours incorrectly predicts that the question 
is irrelevant and refuses to answer a legitimate question. Users
seem more tolerant to mistakes in relevance prediction 
than VQA.

\vspace{\sectionReduceTop}
\section{Conclusion}
\vspace{\sectionReduceBot}
\label{sec:discussions_conclusion}
 
We introduced the novel problem of identifying irrelevant (\ie non-visual or visual false-premise) questions for VQA.
Our proposed models significantly outperform strong baselines on both tasks.
A VQA agent 
that utilizes our detector and refuses to answer certain questions 
significantly outperforms a baseline (that answers all questions) 
in human studies. Such an agent 
is perceived as more intelligent, reasonable, and human-like. %

There are several directions for future work. One possibility includes identifying the premise entailed in a question, as opposed to just stating true- or false-premise. Another is determining what external knowledge is needed to answer non-visual questions.

Our system can be further augmented to communicate to users what the assumed premise of the question is that is not satisfied by the image, \eg 
respond to \emph{``What is the woman wearing?''} for an image of a cat by saying 
\emph{``There is no woman.''}

\textbf{Acknowledgements.}  
We thank Lucy Vanderwende for
helpful suggestions and discussions. We also thank the anonymous reviewers for their helpful comments. This work was
supported in part by the following: National Science
Foundation CAREER awards to DB and DP,
Alfred P. Sloan Fellowship,
Army Research Office YIP awards to DB and DP,
ICTAS Junior Faculty awards to DB and DP, Army
Research Lab grant W911NF-15-2-0080 to DP and
DB, Office of Naval Research grant N00014-14-1-
0679 to DB, Paul G. Allen Family Foundation Allen
Distinguished Investigator award to DP, Google Faculty
Research award to DP and DB, AWS in Education
Research grant to DB, and NVIDIA GPU donation
to DB.
}

\clearpage

\appendix

\renewcommand{\thesection}{Appendix \Roman{section}} 

\section*{Appendix Overview} % Replace with your title

\vspace{-5pt}

In this appendix, we provide the following: 

\begin{compactitem}
\item Additional details for \rulebased (the visual vs. non-visual question detection baseline).
\item Qualitative results.
\item Results with other methods of feature extraction for our question-caption similarity and question-question similarity approaches used 
for true- vs. false-premise question detection.
\item Implementation details for training the models.
\end{compactitem}

\begin{table*}[ht!]
\centering
\begin{tabular}{ccccccc}
\toprule
& & \multicolumn{2}{c}{\textbf{True-Premise}}
&
\multicolumn{2}{c}{\textbf{False-Premise}}
& \\
\cmidrule[0.75pt](l){3-4}
\cmidrule[0.75pt](l){5-6}
& & Recall & Precision & Recall & Precision & \textbf{Norm Acc.} \\
\midrule
\multicolumn{2}{c}{\entropy} & 68.07 & 28.28 & 51.25 & 85.05 & 59.66 \\
\multicolumn{2}{c}{\captionmodelscores} & 64.73 & 25.23 & 50.09 & 84.51 & 57.41 \\
\multicolumn{2}{c}{\vqamlp} & 57.38 & 36.13 & 71.01 & 85.62 & 64.19 \\
\midrule
\multirow{3}{*}{\textbf{\capqsim}}
& \bow & 70.48 & 40.19 & 69.91 & 90.46 & 70.19 \\
& \avgwtov & 69.88 & \textbf{48.81} & \textbf{78.35} & 91.24 & 74.12 \\
& \lstmwtov & 72.37 & 46.08 & 76.60 & 91.55 & 74.48 \\
\midrule
\multirow{3}{*}{\textbf{\qqsim}} 
& \bow & 68.05 & 44.00 & 75.79 & 90.28 & 71.92 \\
& \avgwtov & \textbf{74.62} & 46.51 & 74.77 & \textbf{92.27} & \textbf{74.69} \\
& \lstmwtov & 74.25 & 44.78 & 74.90 & 91.93 & 74.58 \\
\bottomrule
\end{tabular}
\caption{Results for true- vs. false-premise question detection, which are averaged over 40 random train/test splits.}
\label{tab:capsimilarity}
\end{table*}

\begin{figure*}
\includegraphics[width=0.95\textwidth,height=10cm]{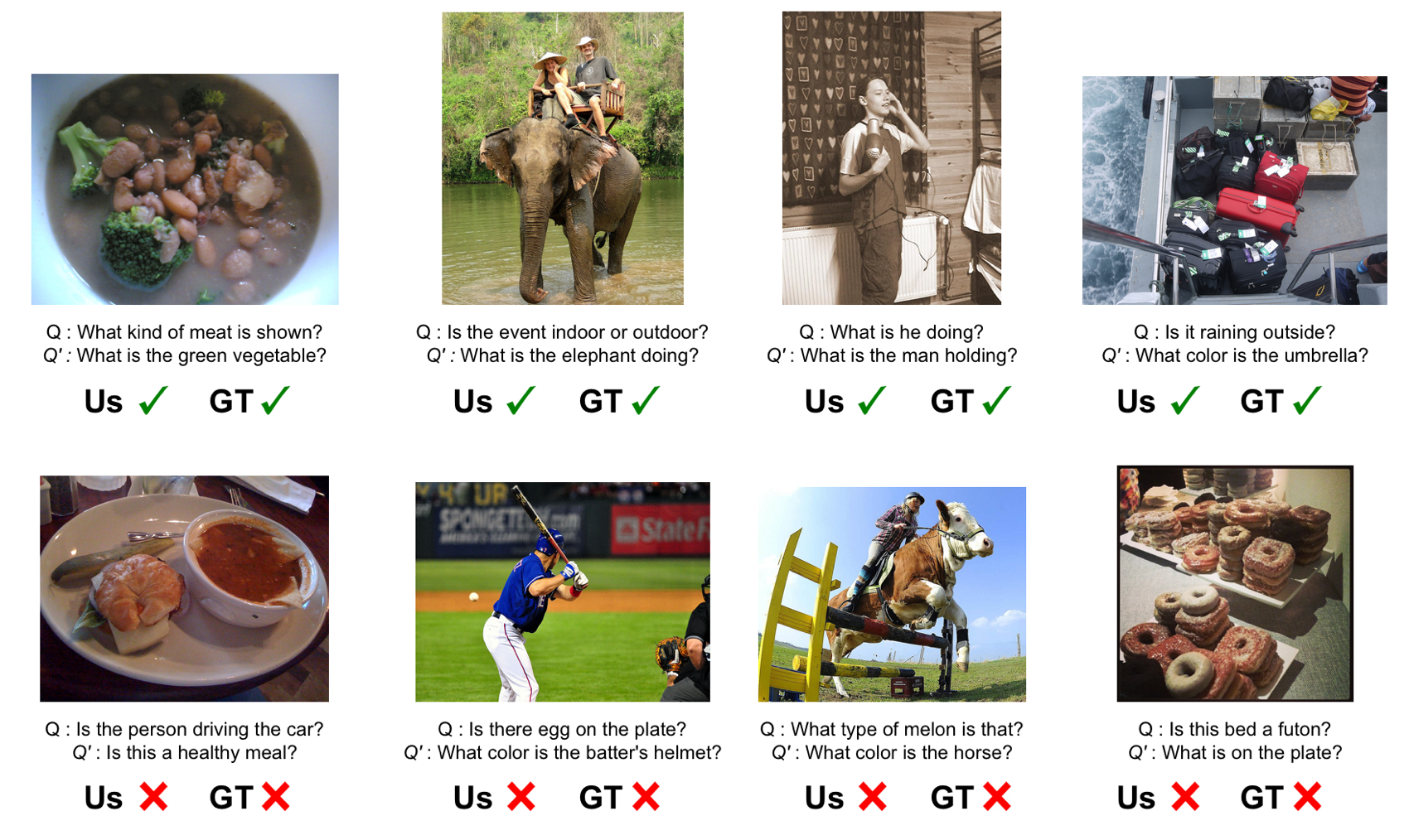}
\caption{Success Cases: The first row illustrates examples that our model thought were true-premise, and were also labeled so by humans. The second row shows success cases for false-premise detection.}
\label{fig:succ}
\end{figure*}

\begin{figure*}
\includegraphics[width=0.95\textwidth, height=10cm]{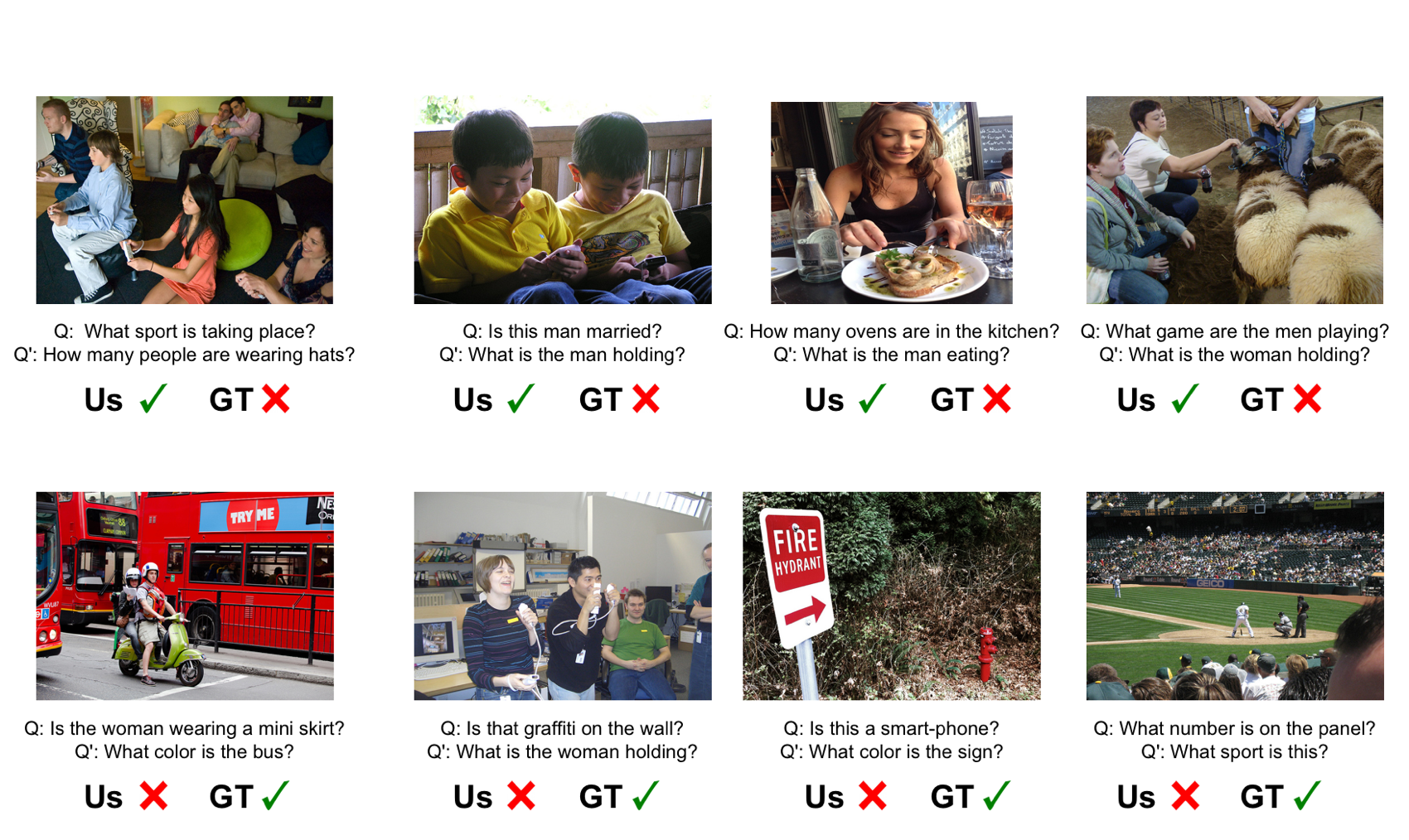}
\caption{Failure Cases: The first row illustrates examples that our model thought was true-premise, but were actually labeled as false-premise by humans. Vice versa in the second row.}
\label{fig:fail}
\end{figure*}

\section{Rule-based Visual vs.~Non-Visual Classification}

Section 4.1 in the main document describes \rulebased, a 
hand-crafted rule-based approach to detect 
non-visual questions.  Rules were added to make this baseline as strong as possible, where some rules take precedence over others. We list a few examples:% to resolve %We proceed over the rules in a particular  

\begin{compactitem}
\item If there is a plural noun, without a determiner before it, followed by a verb 
(\eg \emph{``Do dogs fly?''}),
the question is non-visual. 
\item If there is a determiner followed by a noun (\eg \emph{``Do dogs fly in this picture?''}), the question is visual.
\item If there is a personal or possessive pronoun before a noun (\eg \emph{``What color is his umbrella?''}), the question is visual.
\item We use a list of words that frequently occur in the non-visual questions but infrequently 
in visual questions. 
These include words such as: `God', `Life', `meaning', and `universe'. If any words from this list are present in the question, the question is classified as non-visual.
\end{compactitem}

\section{Qualitative Results}

Here we provide qualitative results for our visual vs. non-visual question detection experiment, and our true- vs. false-premise question detection experiment.

\subsection{Visual vs. Non-visual detection}

Here are some examples of non-visual questions correctly detected by \lstm:
\begin{compactitem}
\item ``Who is the president of the United States?''
\item ``If God exists, why is there so much evil in the world?''
\item ``What is the national anthem of Great Britain?''
\item ``Is soccer popular in the United States?''
\end{compactitem}

Here are some non-visual questions that \rulebased failed on, but that were correctly identified as non-visual by \lstm:
\begin{compactitem}
\item ``What color is Spock's blood?''
\item ``Who was the first person to fly across the channel?''
\end{compactitem}

Here are some visual questions correctly classified by \lstm, but incorrectly classified by \rulebased:
\begin{compactitem}
\item ``Where is the body of water?''
\item ``What color are the glass items?''
\item ``What is there to sit on?''
\item ``How many pillows are pictured?''
\end{compactitem}

\subsection{True- vs False- Premise Detection}
Figures \ref{fig:succ} and \ref{fig:fail} show success and failure cases for true- vs. false- premise question detection using \qqsim. Note that in the success cases, contextual and semantic similarity was learned even when the words in the question generated by the captioning model (Q') were different from the input question (Q). 

\section{Performance of Other Features}

We explored three choices for feature extraction of the questions and captions: 

\begin{compactenum}
\item \bow. We test a bag-of-words approach with a vocabulary size of 9,952 words to represent questions and captions, where we train an MLP to predict whether the question is relevant or not.
The representation is built by setting a value of 1 in the features at the words that are present in either the question or the caption and a 2 when the word is present in both. This means each question-caption pair is represented by a 9,952-dim (vocab length) vector. The MLP used on top of \bow is a 5-layer MLP with 30, 20 and 10 hidden units respectively.

\item \avgwtov. We extract word2vec \cite{mikolov2013distributed} features for the question and captions' words, compute the average of the features separately for the question and caption and then concatenate them.  Similar to \bow, we train a 5-layer MLP with 200, 150 and 80 hidden units, respectively.

\item \lstmwtov. These are the features we used in the main paper. The LSTM has 40 hidden units using a 4-layer MLP with 40 and 20 hidden units respectively.
\end{compactenum}

Table \ref{tab:capsimilarity} shows a comparison of the performance in recall, precision and normalized accuracy, where we have averaged over 40 random train/test splits.

\section{Training Implementation Details}
For training \bow, \avgwtov, \lstmwtov and \vqamlp, we use the Keras Deep learning Library \cite{chollet2015keras} for Python. For pre-training the question and caption generation models from scratch, we use the Torch Deep Learning Library \cite{torch}. We use \emph{rmsprop} as the optimization algorithm (with a learning rate of 0.001) for \lstmwtov, and \emph{adadelta} for \bow and \avgwtov (initialized with a learning rate of 1). For all our models, we use a gaussian random weights initialization and no momentum.

\bibliographystyle{emnlp2016}
\bibliography{emnlp2016,gordon}

\end{document}